\documentclass{article}
\usepackage{arxiv}

\usepackage[utf8]{inputenc} % allow utf-8 input
\usepackage[T1]{fontenc}    % use 8-bit T1 fonts
\usepackage[hidelinks]{hyperref}   % hyperlinks
\usepackage{url}            % simple URL typesetting
\usepackage{booktabs}       % professional-quality tables
\usepackage{amsfonts}       % blackboard math symbols
\usepackage{nicefrac}       % compact symbols for 1/2, etc.
\usepackage{microtype}      % microtypography
\usepackage{graphicx}
\usepackage{amsmath}
\usepackage[table]{xcolor}
\usepackage[warn]{textcomp}
\usepackage{epsfig}
\usepackage{verbatim}
\usepackage{multirow}
\usepackage{txfonts}
\usepackage{float}

\title{StressGAN: A Generative Deep Learning Model for 2D Stress Distribution Prediction}

\author{
  Haoliang Jiang \\
  School of Computer Science\\
  Georgia Institute of Technology\\
  Atlanta, GA 30332\\
  \texttt{hjiang321@gatech.edu} \\
  \And
  Zhenguo Nie \\
  Department of Mechanical Engineering\\
  Carnegie Mellon University\\
  Pittsburgh, PA 15213\\
  \texttt{zhenguon@andrew.cmu.edu} \\
  \And
  Roselyn Yeo \\
  Department of Mechanical Engineering\\
  Carnegie Mellon University\\
  Pittsburgh, PA 15213\\
  \texttt{RYEO005@e.ntu.edu.sg} \\
  \And
  Amir Barati Farimani \\
  Department of Mechanical Engineering\\
  Carnegie Mellon University\\
  Pittsburgh, PA 15213\\
  \texttt{barati@cmu.edu} \\
  \And
  Levent Burak Kara\thanks{Address all correspondence to this author.} \\
  Department of Mechanical Engineering\\
  Carnegie Mellon University\\
  Pittsburgh, PA 15213\\
  \texttt{lkara@cmu.edu} \\
}
\begin{document}
\maketitle

%%%%%%%%%%%%%%%%%%%%%%%%%%%%%%%%%%%%%%%%%%%%%%%%%%%%%%%%%%%%%%%%%%%%%%
\begin{abstract}
Using deep learning to analyze mechanical stress distributions has been gaining interest with the demand for fast stress analysis methods. Deep learning approaches have achieved excellent outcomes when utilized to speed up stress computation and learn the physics without prior knowledge of underlying equations.  However, most studies restrict the variation of geometry or boundary conditions, making these methods difficult to be generalized  to unseen configurations. We propose a conditional generative adversarial network (cGAN) model for predicting 2D von Mises stress distributions in solid  structures. The cGAN learns to generate stress distributions conditioned by geometries, load, and boundary conditions through a two-player minimax game between two neural networks with no prior knowledge. By evaluating the generative network on two stress distribution datasets under multiple metrics, we demonstrate that our model can predict more accurate high-resolution stress distributions than a baseline convolutional neural network model, given various and complex cases of geometry, load and boundary conditions. 
%The full implementation (based on Tensorflow) and the trained networks are available at \url{https://github.com/zhenguonie/TopologyGAN}.

%\noindent Keywords: TopologyGAN, Deep Learning, Topology Optimization, U-SE-ResNet, Physical Fields
\end{abstract}

%%%%%%%%%%%%%%%%%%%%%%%%%%%%%%%%%%%%%%%%%%%%%%%%%%%%%%%%%%%%%%%%%%%%%%
\section{Introduction}
Structural stress analysis is a critically important foundational tool in many disciplines including mechanical engineering, material science, and civil engineering. It is used for predicting the stress distribution and the possibility of structural failure when the structure is subject to the applied load and boundary conditions \cite{VonMisesTOPO, miseschem, misesbone}. Finite Element Analysis (FEA) is commonly used to discretize the domain and to solve the governing partial differential equations \cite{appliedFEA, FEAconcepts, SF_DL, Mises1913}. Traditional methods provide high fidelity solutions but require the solution of large linear systems which can be computationally prohibitive. With the demand for fast and accurate structural analysis in generative design, topology optimization technologies and online manufacturing monitoring, increasing the computational speed for stress analysis has become a focus of interest. 

To achieve fast mechanics analysis, many prior works have focused on deep learning techniques to help compute computational engineering problems \cite{deeplearning}. Several approaches of accelerating mechanical stress analysis by deep learning methods have been carried out and achieved excellent outcomes in terms of computational speed and accuracy \cite{surrogate_FEA, 3Dtrusses, 3Dprint, SF_DL, residual_pipe, shear}. These studies utilize deep learning models to predict residual stress, shear stress, maximum von Mises stress or distributions of the stress tensor. Once trained on large datasets, these approaches are able to generate accurate stress predictions. However, most previous work restricts the geometry or the boundary conditions that are applied, making the models difficult to be generalized to new problems. 

In this work, we propose a conditional generative adversarial network we call StressGAN for stress distribution prediction. 
StressGAN takes as input arbitrary geometries, load and boundary conditions in the form of different input channels and predicts the von Mises stress distribution in an end-to-end fashion. A distinguishing feature of our approach is that we utilize a generative adversarial network instead of an autoencoder as our learning algorithm. 

We evaluate StressGAN on two datasets: a fine-mesh multiple-structure dataset introduced by this work and a coarse-mesh cantilever beam dataset used in \cite{SF_DL}. The fine-mesh dataset contains 38,400 problem samples modeled as $128 \times 128$ meshes. Unlike the coarse-mesh dataset with identical shape, boundary and load conditions, the fine-mesh dataset includes ten patterns of load positions and eight patterns of boundary conditions. To explore the performance of StressGAN under different scenarios, we design two types of experiments 
% based on characteristics of datasets
: training and evaluating the network on entire dataset and training and evaluating the network on datasets with conditions of different categories (generalization experiments). 
As a result, StressGAN outperforms a selected baseline model, StressNet (SRN), proposed in \cite{SF_DL}, on the fine-mesh dataset and generates reasonably accurate results on the coarse-mesh dataset. Furthermore, StressGAN generates relatively accurate stress distributions for most test cases in the generalization experiments with sparse training dataset.

%%%%%%%%%%%%%%%%%%%%%%%%%%%%%%%%%%%%%%%%%%%%%%%%%%%%%%%
\section{Related Work}
We focus our review on finite element analysis, then on studies that focus on  deep learning methods with emphasis on their applications in stress estimation and generative adversarial networks with emphasis on their  applications in computational engineering.
\subsection{Finite element analysis for stress computation}
% Solid mechanics analysis of a given geometry requires to solve the related partial differential equations. Finite element analysis (FEA) is a mature computational tool to solve this problem. It simplifies the structure by breaking it down into large number of finite elements and analyzes the mechanical displacements based on the given conditions by building up an algebraic equation:
Typical linear finite element analysis (FEA) for stress calculations involve:
\begin{equation}
KQ=F
\end{equation}
Where $K$ denotes a global stiffness matrix; $F$ denotes a vector describing the applied load at each node; $Q$ denotes the displacement. $K$ is composed of elemental stiffness matrices $k_e$  for each element:
\begin{equation}
k_e = A_e B^T D B
\end{equation}

\noindent where $B$ is the strain/displacement matrix; $D$ is the stress/strain matrix; $A_e$ is the area of the element. $B$ and $D$ are determined by material properties and mesh geometry. Then the local stiffness matrix $k_e$ will be added into the global stiffness matrix.  The  displacement boundary conditions are encoded into the global stiffness matrix $K$ by operating on the corresponding rows and columns. Various direct factorization based or iterative solvers exist for the solution of $Q$.

After computing the global displacement using equation (1), the nodal displacement $q$ of each element, followed by the stress tensor of each element:
\begin{equation}
\sigma = DBq
\end{equation}
Where $\sigma$ denotes the tensor of an element. Then, the von Mises Stress of each element is computed using the 2-D von Mises Stress form:
\begin{equation}
\sigma_{vm} = \sqrt{\sigma_{x}^2 + \sigma_{y}^2 - \sigma_{x}\sigma_{y}+3\tau_{xy}^2}
\end{equation}

\noindent where $\sigma_{vm}$ is von Mises Stress; $\sigma_{x}$, $\sigma_{y}$, $\tau_{xy}$ are the normal and shear stress components. 

\subsection{Deep learning in mechanical stress analysis}
Most of the early attempts to use deep learning in speeding up mechanical stress analysis focus on integrating the models in a FEA software. These models are to solve some auxiliary tasks including updating FEA model \cite{FEA_modelUpdate1, FEA_modelUpdate2}, checking plausibility of a FE simulation \cite{NN_plausibilityCheck}, modeling the constitutive relation of a material  \cite{NN_FAE} and optimizing the numerical quadrature in the element stiffness matrix on a per element basis \cite{kmatrix} . These works alleviate the complexity of FEA software to some extent. However, our approach could be used as a surrogate model to a FEA software. It avoids the computation bottlenecks in a FEA software and its computation cost could be controlled by modifying the architecture. 

Deep learning methods are proposed as surrogate models to approximate residual stress in girth welded pipes \cite{residual_pipe}, shear stress in circular channels \cite{shear} or stress in 3D trusses \cite{3Dtrusses}. These methods use manually assigned features to represent a fixed geometry or a part of the geometry. The deep learning models will estimate a stress value based on the input parameters. In our work, the deep learning method learns to filter useful features and generates a representation for each combination of the geometry, external load and boundary condition. A decoder follows the data representation and predicts a stress distribution on the geometry.

Liang et al. \cite{surrogate_FEA} have developed an image-to-image deep learning framework as an alternative to predicting aortic wall stress distributions by expanding aortic walls into a topologically equivalent rectangle. Khadilkar et al. \cite{3Dprint} propose a two-stream deep learning framework to predict stress fields in each of the 3D printing process. The network encodes 2D shapes of each layer and the point clouds of 3D models based on a CNN architecture and a PointNet \cite{PointNet}. Madani et al. \cite{vessel} propose a transfer learning model to predict the value and position of the maximum von Mises stress on arterial walls in atherosclerosis. Our model also use an image-to-image translation model to estimate the stress distribution. However, we utilize image-based data representation on both the geometry and the input conditions. Thus, our model can be used to analyze arbitrary 2D stress distribution cases after proper training.

Most related to our work, Nie et al. \cite{SF_DL} propose an end-to-end convolutional neural network called StressNet to predict 2D stress distributions given multi-channel data representations of geometry, load and boundary conditions of cantilever beams. The network contains three downsampling convolutional layers, five Squeeze-and-Excitation ResNet (SE-RES) blocks \cite{resnet, senet} and three upsampling convolutional layers. Each SE-RES block is composed of two convolutional layers and a SE block which utilizes a global pooling and two fully connected layers to learn extra weights for each channel. Skip connections are used in each block. 9x9 kernels are used in the first and last convolutional layer and 3x3 kernels are used in all remaining convolutional layers. A dataset composed of 120,960 cases of cantilever beams modeled using $32 \times 32$ meshes is generated by a linear FEA software to train and evaluate the network. In our work, we aim at analyzing high-resolution cases and use an adversarial learning scenario additionally to capture features in stress distributions. More importantly, all previous work of deep learning methods in stress prediction focus on specific application cases lacking variety in geometry, external load and boundary conditions. Moreover, through testing our model by geometries or conditions from unseen domain, we show the potential of our deep learning model as a transfer learning model for stress field predictions.

\subsection{Generative adversarial networks}
\begin{figure}[ht]
	\centering
	\includegraphics[width=0.5\linewidth]{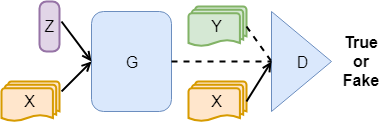}
	\caption{Framework of cGAN. The generator G and discriminator D are conditioned on information $X$. A latent vector $Z$ and $X$ compose the input to G. D learns to tell whether its input regarding $X$ is from real samples $Y$ or output of G.
	}
	\label{fig:cGAN}
\end{figure}

GANs are an example of generative models. They model the training of a generative network as a two-player minimax game where a generator $G$ is trained to learn a distribution $f$ with a discriminator $D$ \cite{GAN}. Both of them represent a differentiable function contolled by the learned parameters. In a conventional GAN, the generator $G$ learns to map a vector sampled from a latent space $z \sim p_z(z)$ to the space of ground truth samples. In the meantime, the discriminator $D$ learns to map a sample to a probability that predicts if the presented sample is real or fake. The Nash equilibrium in training is that the generator forms the same distribution as the training data and the discriminator output 0.5 for all input data \cite{goodfellow2016nips}.

cGAN is built upon the learning algorithm of GAN and has been widely used to date \cite{pix2pix, SRGAN, liu2017unsupervised, ROCGAN}. cGAN develops a method to control the mapping from input to output by conditioning the standard generator $G$ and discriminator $D$ on extra information. Figure \ref{fig:cGAN} demonstrates the framework of cGAN. Further, Isola et al. \cite{pix2pix} propose a similar network for image-based task such as image-to-image translation. In the comparisons against networks plainly using MAE as a loss function, it shows the superiority of using cGAN framework in image-based tasks. Radford et al. \cite{DCGAN} reinforce GAN's training stability by using all convolutional net \cite{ACN}, ReLU and LeakyReLU activations and batch-normalization layers. 

Farimani et al. \cite{Amir_phaseSegregatoin, Amir_Transport} propose a cGAN architecture based on the network proposed by Isola et al. \cite{pix2pix} to learn models of steady state heat conduction, incompressible fluid flow, and phase segregation. S. Lee et al. use GAN in the prediction of unsteady flow over a circular cylinder with various Reynolds numbers \cite{Lee_cylinder}. Paganini et al. \cite{colagan} use a revised DCGAN is developed to model electromagnetic showers in a longitudinally segmented calorimeter. The deep learning method speeds up the calculations by more than 100 times. K. Enomoto et al. also utilize a DCGAN architecture for cloud removal in climate images \cite{cloudRemove}. In the field of astronomy, GANs are used to generate images of galaxies \cite{ETH_galaxy, CMU_dackEnergy} and 2D mass distributions \cite{CosmoGAN}.  In our work, we use the architecture and learning algorithm introduced by Radford et al. \cite{DCGAN} and Isola et al. \cite{pix2pix} to build our neural network for stress field predictions cross varying input geometries and boundary conditions.

%%%%%%%%%%%%%%%%%%%%%%%%%%%%%%%%%%%%%%%%%%%%%%%%%%%%%%%
\section{Technical Approach}
\subsection{Neural Network Architecture}
\begin{figure}[ht]
	\centering
	\includegraphics[width=0.5\linewidth]{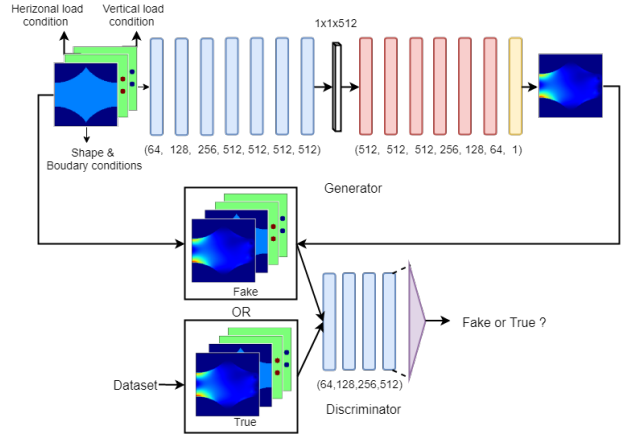}
	\caption{Architecture of StressGAN. The generator (top) and the discriminator (bottom) are constructed with downsampling blocks (blue) and upsampling blocks (red). For the last upsampling block of the generator (yellow), we remove the ReLU activation. 
	The numbers indicate channel dimensions of the output of each blocks. 
	%When the network is trained and tested using rough-mesh dataset, we remove two blocks close to the bottleneck to keep the representation of input conditions as a $512$ feature vector. 
	The purple triangle means a reshape layer followed by a linear layer and a Sigmoid activation.
	}
	\label{fig:StressGAN}
\end{figure}

The architecture of StressGAN is shown in figure \ref{fig:StressGAN}. We design the generator as an encoder-decoder network which generates a feature vector with a size of $512$ in the bottleneck. The input of the generator is a case of conditions and geometry modeled by $m \times m$ meshes. Three $m \times m$ resolution images are used to represent geometry, boundary conditions and the applied load. To increase data intensity, we represent geometry and boundary conditions on one image. We use numbers 0, 1, 2, 3, 4 in geometries to represent various boundary conditions, where 0 is void, 1 means free solid node, 2 means node affixed horizontally, 3 means node affixed vertically, 4 means node affixed on both directions. The remaining two images record magnitudes of vertical or horizontal loads in the corresponding pixel. The output of the generator is a $m \times m$ mesh describing the von Mises stress distributions.
% It will be $N \times  N$ for stress field prediction; $m \times m \times 2$ for displacement prediction because we use two channels of image to represent displacement on vertical and horizontal directions. 
The encoder is comprised of $\mathrm{log_2}(m)$ downsampling blocks with a convolutional layer, a batch normalization layer and a LeakyReLU layer. Similarly, the decoder is comprised of $\mathrm{log_2}(m)$ upsampling blocks with a deconvolutional layer, a batch normalization layer and a ReLU layer. When the network is trained and tested using the coarse-mesh dataset, we remove four blocks close to the bottleneck to keep the bottleneck representation of input conditions as a $512$ feature vector. We remove the ReLU layer in the last upsampling block with the consideration that mechanics analysis results other than von Mises Stress might contain negative values. The convolutional layers and deconvolutional layers both have kernel sizes as $5 \times 5$ and stride size as 2. 

For the discriminator, we adopt a downsampling structure. The input is a stress distribution case described by four $m \times m$ images including the fake or ground truth sample stress distribution and its conditions. The architecture of the discriminator is fixed when experimented on different datasets. It outputs a probability value which describes whether the input analysis result is true regarding the conditions and geometry. Four downsampling blocks are followed by a reshape layer, a fully connected layer, and a Sigmoid activation.
% The architecture will be introduced using two paragraphs: generator and discriminator.(Mention other architectures and give the stats in the supplementary information part, especially w/o unet and bottleneck size)
\subsection{Loss function and metrics}
\textbf{Loss function} Our loss function consists of an L2 distance loss and a cGAN objective function:
\begin{equation}
\mathrm{L_{L2}}(G) = \mathrm{E_{x,y}}[||y-\mathrm{G}(x)||_2]
\end{equation}

\noindent where $y$ is ground truth stress distributions; $x$ stands for conditions and geometries, $G$ denotes the generator. Previous work has shown that utilizing L2 distance (MSE) to train networks for predicting stress distributions works well. Thus, we use L2 distance as a loss in StressGAN's loss function.

The loss function of cGAN used in our model can be expressed as:
\begin{equation}
\begin{aligned}
\mathrm{min_Gmax_DV}(G, D) = \mathrm{E_{x,y}}[\mathrm{log(D}(x, y))] + \\\mathrm{E_{x}}[\mathrm{log(1-D(}x,\mathrm{G}(x)))]
\end{aligned}
\end{equation}

cGAN loss function shows the adversarial relationship between the generator $G$ and the discriminator $D$. Note that in our cases where the network should output a particular analysis result given the conditions and a geometry, we eliminate the Gaussian noise vector $z$ which is usually an input of the generator to add more variation to the  output.

The final loss is:
\begin{equation}
\mathrm{min_G} \mathrm{max_DV}(G,D) + \lambda \mathrm{L_{L2}}(G)
\end{equation}

\noindent where a hyperparameter $\lambda$ is to balance the loss function following \cite{pix2pix, SRResnet}. 

\textbf{Metrics} In addition to MSE, four metrics are introduced to assess the performance of StressGAN: mean absolute error (MAE), percentage mean absolute error (PMAE), peak stress absolute error (PAE) and percentage peak stress absolute error (PPAE). These four metrics, whether related to MSE or not, are not an explicit goal of minimizing MSE. Using these four metrics, we can provide an  assessment of stress prediction qualities.

Using MAE and a normalized version of MAE, PMAE, helps evaluate the overall quality of a predicted stress distribution. MAE is defined as:

\begin{equation}
\mathrm{MAE} = \frac{1}{n} \sum_{j=1}^{n} |y_j - \hat{y_j}|
\end{equation}

\noindent where $y_j$ is the value at element j in a ground truth sample; $\hat{y_j}$ is the estimated value at element j; n denotes the number of elements of samples. PMAE is defined as:

\begin{equation}
\mathrm{PMAE} = \frac{\mathrm{MAE}}{\mathrm{max\{}Y\} - \mathrm{min\{}Y\}} \times 100 \%
\end{equation}

\noindent where $Y$ denotes a set of all ground truth stress values in a case; $\mathrm{max\{}Y\}$ is the maximum value in a set of ground truth stress values $Y$; $\mathrm{min\{}Y\}$ is the minimum value in a set of ground truth stress values $Y$.

PAE and PPAE measure the accuracy of the most considerable stress value in a predicted stress distribution which is the most critical local value of stress distributions in engineering applications. PAE is defined as:

\begin{equation}
\mathrm{PAE}\ =|\mathrm{max\{}Y\} - \mathrm{max\{}\hat{Y}\}|
\end{equation}

\noindent where $\hat{Y}$ is the set of all predicted stress values in a case; $\mathrm{max\{}\hat{Y}\}$ is the maximum value in a set of all predicted stress values $\hat{Y}$.

PPAE is defined as:
\begin{equation}
\mathrm{PPAE}\ = \frac{\mathrm{PAE}}{\mathrm{max}\{Y\}} \times 100\%
\end{equation}

%%%%%%%%%%%%%%%%%%%%%%%%%%%%%%%%%%%%%%%%%%%%%%%%%%%%%%%
\section{Experiments}
\subsection{Dataset and implementation}
\begin{figure}[ht]
	\centering
  \includegraphics[trim = 0in 0in 0in 0in, clip, width=0.47\textwidth]{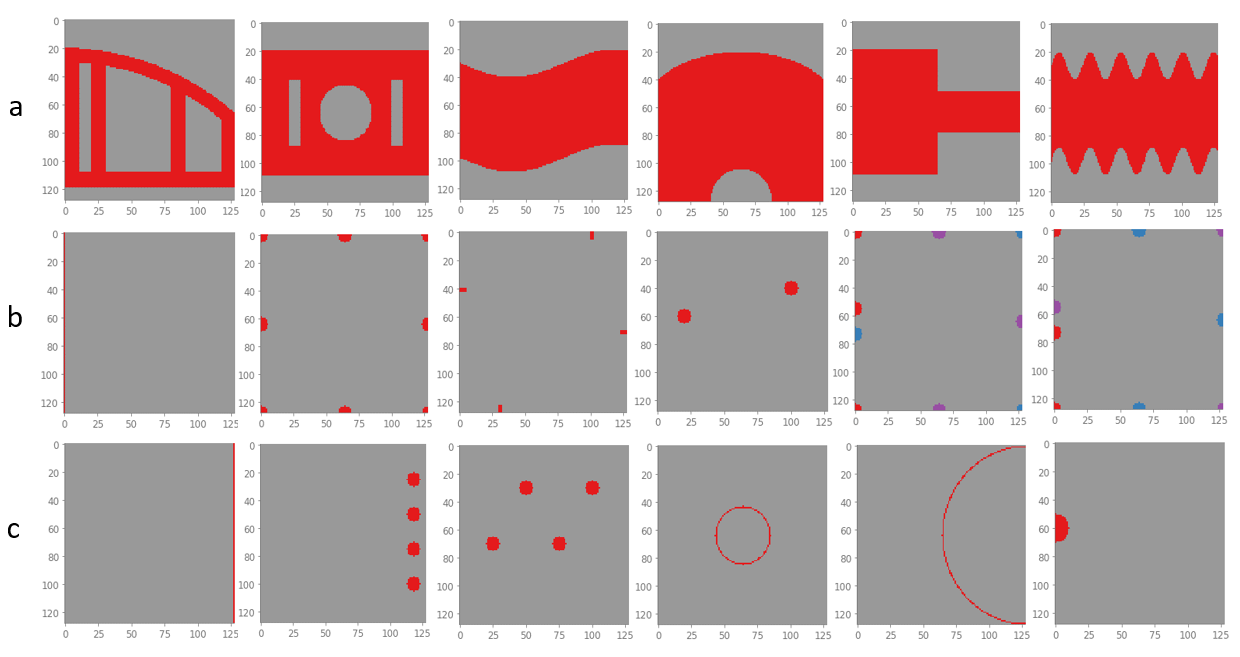}
	\caption{Data samples in fine-mesh dataset. a. Geometries.  b. Boundary conditions.  c. Load positions.
	}
	\label{fig:finemesh}
\end{figure}

\textbf{Fine-mesh multiple structure dataset} To provide a broad evaluation of our network's performance, we introduce a stress distribution dataset composed of multiple structures each modeled as a $128 \times 128$ elements. The dataset is generated using A 2D FEM software SolidsPy \cite{solidspy}. All elements in the domain is a 4-node quadrilateral with a size of $1 \times 1$ (mm). Void regions are modeled using a Young’s modulus of infinitesimal value. The dataset contains 60 geometries, ten patterns of load conditions and eight patterns of boundary conditions, in total, 38,400 instances. The shapes, load conditions and boundary conditions are not limited to cantilever beams which are affixed on one end and bearing loads on the other end. Samples of geometry, load and boundary conditions are demonstrated in Figure \ref{fig:finemesh}. Also, for each load pattern, the orientations of the loads can vary from 0 degrees to 315 degrees with a step of 45 degrees. We normalize the load magnitudes in the dataset to reduce the input space since the linear characteristic of homogeneous and isotropic elastic material results in a linear relationship between the loads and the stresses. 

\begin{figure}[h]
	\centering
  \includegraphics[trim = 0in 0in 0in 0in, clip, width=0.45\textwidth]{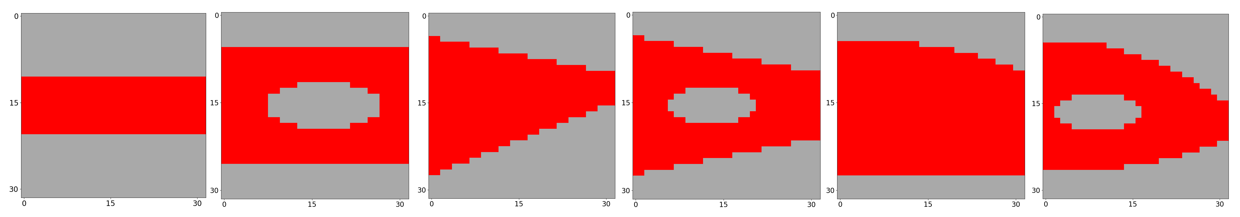}
	\caption{Categories of geometry in coarse-mesh dataset. From left to right: rectangular beam; rectangular beam with a cellular opening; trapezoidal beam; trapezoidal beam with a cellular opening; beam with parabola contours; beam with parabola contours and a cellular opening
	}
	\label{fig:coarsemesh}
\end{figure}

\textbf{Coarse-mesh cantilever beam dataset} The course-mesh stress distribution dataset is proposed by Nie et al. \cite{SF_DL}. The dataset consists of six categories of geometries, in total, 80 geometries. Examples of categories of geometry are shown in Figure \ref{fig:coarsemesh}. Load is applied on the right end of the beam. The left end of the beam is fixed. For each geometry, load orientation changes from 0 degrees to 355 degrees, in 5 degree increments. For each orientation, the load magnitudes varies from 0N to 100N by a step of 5N. In total, the dataset includes 120,960 instances with various shapes, load orientations, and load magnitudes. 
% To improve the data intensity and to save the computational cost, we embed the boundary conditions into the images of geometry. 
% \textbf{Samples from the dataset are shown in the Appendix.}

\textbf{Implementation detail}  We train StressGAN using a learning rate of 0.001 by the Adam optimizer \cite{Adam} with a batch size of 64. We use 1 or 10 as the value of $\lambda$ in StressGAN's loss in the experiments with the fine-mesh dataset and coarse-mesh dataset respectively. The learning rate and batch size are decided by a grid search on potential values. The performance of the model and fitting GPU memory size are main considerations in the grid search. The selection of $\lambda$ is up to the balance between L2 loss and cGAN loss. We aim to keep the losses at the same weight for stabilizing the training process.  In each training epoch, we train the discriminator once and the generator twice to keep the training stable. In all experiments, we use an NVIDIA GeForce GTX 1080Ti GPU. Under our experiment setting, each cases in fine-mesh dataset take approximately 0.003 seconds to analyze.

\subsection{Experiment Design}
\textbf{Entire dataset training and evaluation}
In this experiment, we randomly divide both the fine-mesh dataset and the coarse-mesh dataset into train/test sets of $80\%-20\%$ split respectively. We then train and evaluate StressGAN on the datasets to demonstrate its effectiveness. Additionally, we train and evaluate our baseline model SRN under the same scenario to compare their performances.

\textbf{Generalization training and evaluation} To further study StressGAN's performances in general engineering scenarios such as unseen geometries or unseen applied loads, we set three sub-experiments where training and test sets belonging to different categories of geometry or load orientation. The whole experiment is set based on coarse-mesh dataset since it is easier to separate geometries into semantic categories. In the first and second sub-experiments, we train and evaluate the networks using samples from different geometry categories respectively. In the first sub-experiment, we train the networks with samples in the categories of rectangular beams, trapezoidal beams, rectangular beams with cellular openings and trapezoidal beams with cellular openings and evaluate the networks with beams with a parabola contour. In the second experiment, we train the networks using beams without holes and evaluate the networks using beams with cellular openings. The third sub-experiment is to study how the network performs when trained and evaluated by cases with different load orientations. The load orientations are split up by quadrants. We randomly select loads in three quadrants for training and use loads in the remaining quadrant for testing. We normalize the load magnitudes in all training and test datasets, which reduces the size of all training datasets to less than 5000 samples.

%%%%%%%%%%%%%%%%%%%%%%%%%%%%%%%%%%%%%%%%%%%%%%%%%%%%%%%
\section{Results and Discussions}
\subsection{ Entire dataset evaluation}

\begin{figure}[ht]
	\centering
  \includegraphics[trim = 0in 0in 0in 0in, clip, width=0.5\textwidth]{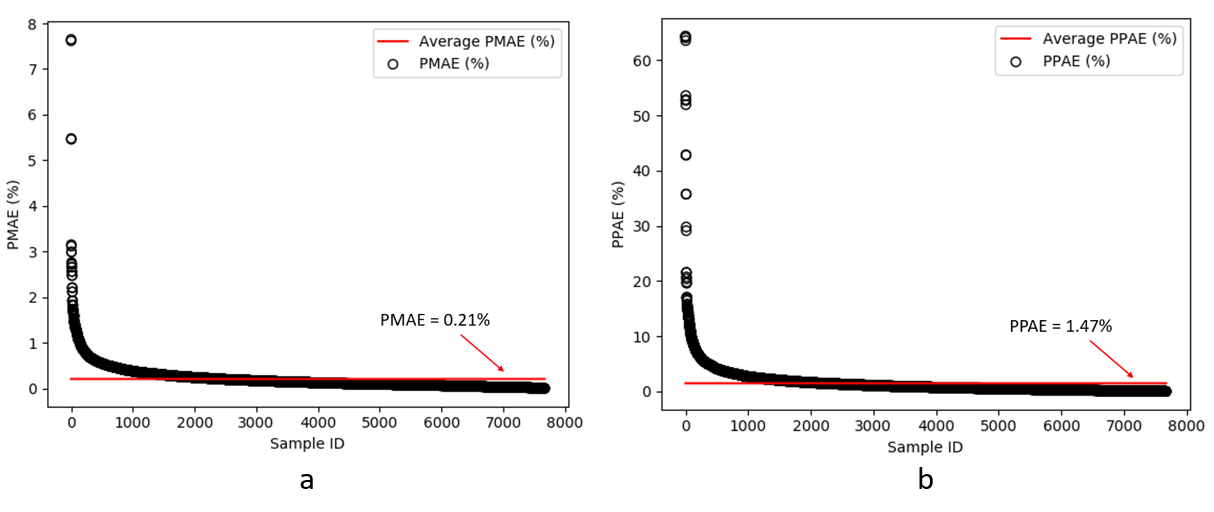}
	\caption{Statistical accuracy of StressGAN on fine mesh dataset. a. PMAE of each sample and average PMAE on the test dataset. b. PPAE of each sample and average PPAE on the test dataset.
	}
	\label{fig:AE_128}
\end{figure}

\begin{figure}[ht]
	\centering
	\includegraphics[width=0.5\linewidth]{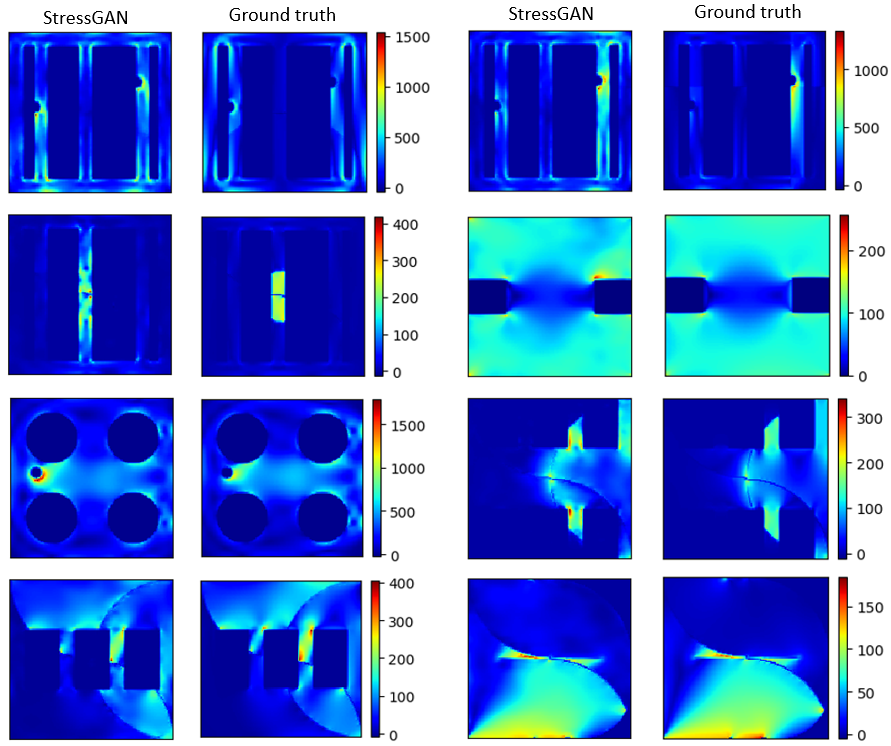}
	\caption{The worst predictions of StressGAN on fine-mesh dataset. 
	%The most inaccurate cases of StressGAN predictions on fine-mesh dataset and the ground truths are shown. 
	%Although these predictions of StressGAN are relatively not accurate, they are still approximately similar to the ground truth stress distributions.
	}
	\label{fig:worstcase}
\end{figure}

\begin{figure}[ht]
	\centering
	\includegraphics[width=0.5\linewidth]{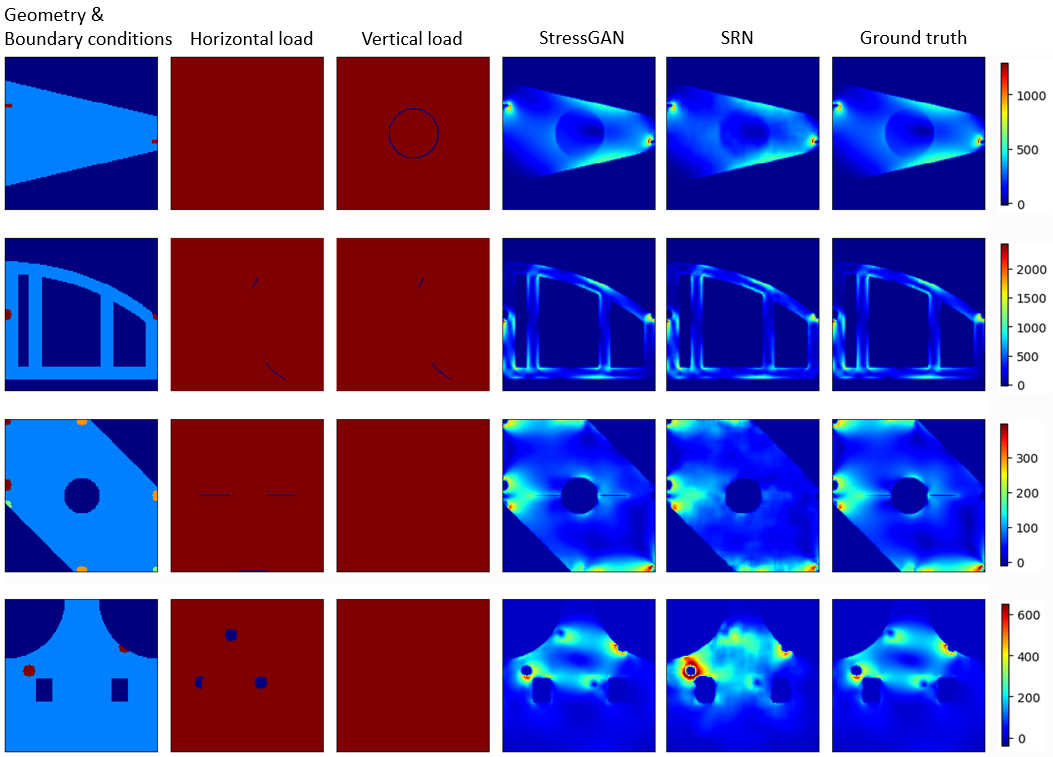}
	\caption{Evaluation of StressGAN and SRN on fine-mesh dataset. Four evaluation cases are shown by each row. From left to right: 1) Geometry (light blue) and boundary conditions (cyan: horizontal, orange: vertical, red: vertical and horizontal); 2) horizontal load positions; 3) vertical load positions; 4)  predictions of StressGAN; 5) predictions of SRN; 6) ground truths. The load magnitudes of each case: 
	1) $\mathrm{q}(x) = 0.0 N/mm^2$, $\mathrm{q}(y) = -88.4 N/mm^2$;
	2) $\mathrm{q}(x) = 125.0 N/mm^2$, $\mathrm{q}(y) = 125.0 N/mm^2$;
	3) $\mathrm{q}(x) = 100.0 N/mm^2$, $\mathrm{q}(y) = 0.0 N/mm^2$;
	4) $\mathrm{q}(x) = -68.8 N/mm^2$, $\mathrm{q}(y) = 0.0 N/mm^2$. (Units: mm-MPa-N) 
	}
	\label{fig:128_combine}
\end{figure}

As stated previously, we train and test our model using the fine-mesh dataset with a split of 80\% - 20\%. Meanwhile, 
we train and test SRN on the same training and testing dataset. The evaluation results of the three networks are shown in Table \ref{tab: 128_stress}. The best performance under each metric are shown in bold. StressGAN attains a PMAE of $0.21\%$ and a PPAE of $1.47\%$, which indicates StressGAN can produce accurate fine-mesh stress distribution given complex cases. The statistical accuracy of StressGAN on the test dataset is shown in Figure \ref{fig:AE_128}. The most inaccurate predictions are shown in Figure \ref{fig:worstcase}. Even with the highest related error rates, these predictions still provide useful information. Table \ref{tab: 128_stress} shows that StressGAN outperforms SRN with a significant margin in all metrics. Figure \ref{fig:128_combine} shows comparisons of the evaluation results of StressGAN and SRN. As shown in the visualizations, the predicted stress distributions of StressGAN are sharper than the predictions of SRN, especially around the edges of the void versus material boundaries. Additionally, StressGAN's predictions of the critical areas are comparatively more accurate.

% \begin{figure}[ht]

% \begin{figure}[!tp]
\begin{figure}[ht]
	\centering
  \includegraphics[trim = 0in 0in 0in 0in, clip, width=0.5\textwidth]{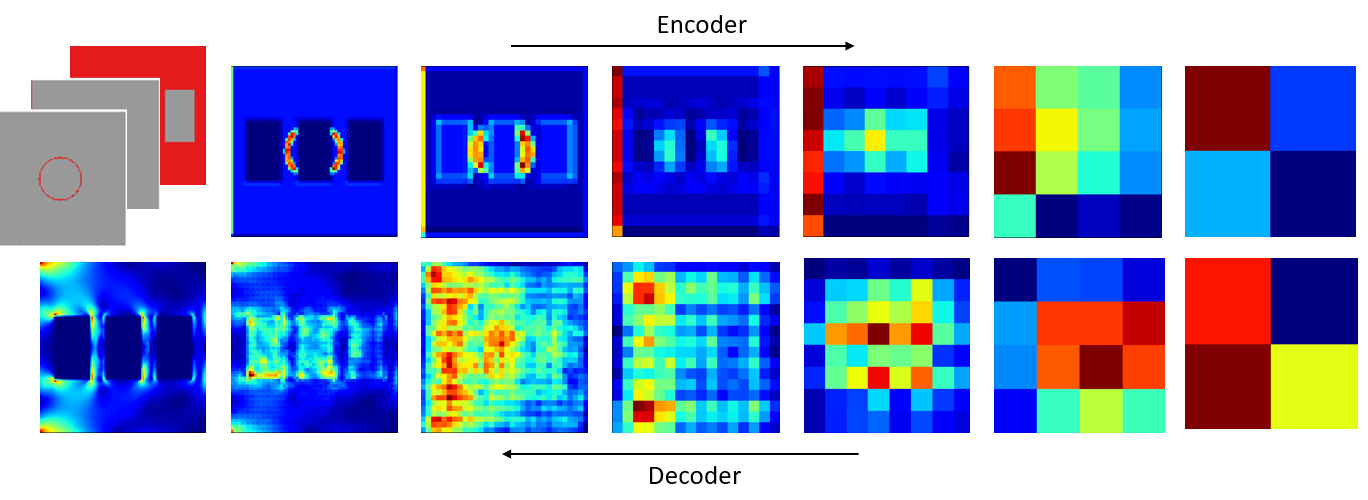}
	\caption{Output activation layers of the generator. The output activation layers of convolutional layers are shown to display the encoding and decoding processes.
	}
	\label{fig:encoder_decoder}
\end{figure}

We also visualize the activation layers of a random sample and test the discrimination of the discriminator. Figure \ref{fig:encoder_decoder} shows the output activation layers of the convolutional layers in the generator. From the figures in the upper row, it can be clearly observed that the positions of boundary conditions and external forces are highlighted by the filters, which demonstrates the ability of the network to capture and transfer input conditions. The figures in the bottom row provides insights into how the network computes the stress distribution based on the encoded information. Ground truths and predictions of test dataset are fed into the discriminator. The average output of the discriminator given the ground truths and  predictions is 0.899 and 0.002, respectively. This shows that the discriminator has learned the implicit features of stress distributions. Even with test data, it is able to distinguish the ground truth distribution from the predicted ones.

\begin{table}[!htb]
\centering
\caption{\label{tab: 128_stress} Quantitative evaluation of StressGAN and SRN with find-mesh dataset.
% The table shows evaluation results of StressGAN and SRN
%which demonstrates the performance of StressGAN in predicting high-resolution and diverse stress distributions. 
The best result under each metric is shown in bold.  (Units: mm-MPa-N)}
% StressGAN outperforms SRN with a significant gap in all metrics.

\begin{tabular}{cccccc}
\hline
Metric    & MSE                                                  & MAE                      & PMAE                          & PAE                     & PPAE                         \\ \hline
StressGAN & \textbf{77.31}                                               & \textbf{1.83} & \textbf{0.21}\% & \textbf{20.29} & \textbf{1.47}\% \\ 
% StressGAN-G & \textbf{75.23} & 2.179                         & 0.25\%                         & 25.33                         & 3.78\%                         \\ \hline
SRN       & 1119.75                                              & 10.88                         & 1.20\%                         & 132.64                         &  19.80\%                         \\ \hline
\end{tabular}
\end{table}

\begin{table}[!htb]
\centering
\caption{\label{tab: Cantilever_beam} Quantitative evaluation of StressGAN and SRN with coarse-mesh dataset. The best performance under each metric is shown in bold. (Units: mm-MPa-N)}
% The table shows the performance of StressGAN and SRN when trained and evaluated using Cantilever Beams Dataset. 
% The best result under each metric is shown in bold. Although SRN outperforms StressGAN in all metrics, the accuracy of StressGAN's predictions is fairly high. 

\begin{tabular}{cccccc}
\hline
Metric          & MSE            & MAE     & PMAE            & PAE     & PPAE             \\ \hline
StressGAN & 1.08 & 0.60 & 0.59\% & 2.17          & 2.11\%          \\ 
SRN       & \textbf{0.15}          & \textbf{0.20}          & \textbf{0.15\%}          & \textbf{0.50} & \textbf{0.37\%} \\ \hline
\end{tabular}
\end{table}

\begin{figure}[ht]
	\centering
	\includegraphics[width=0.45\linewidth]{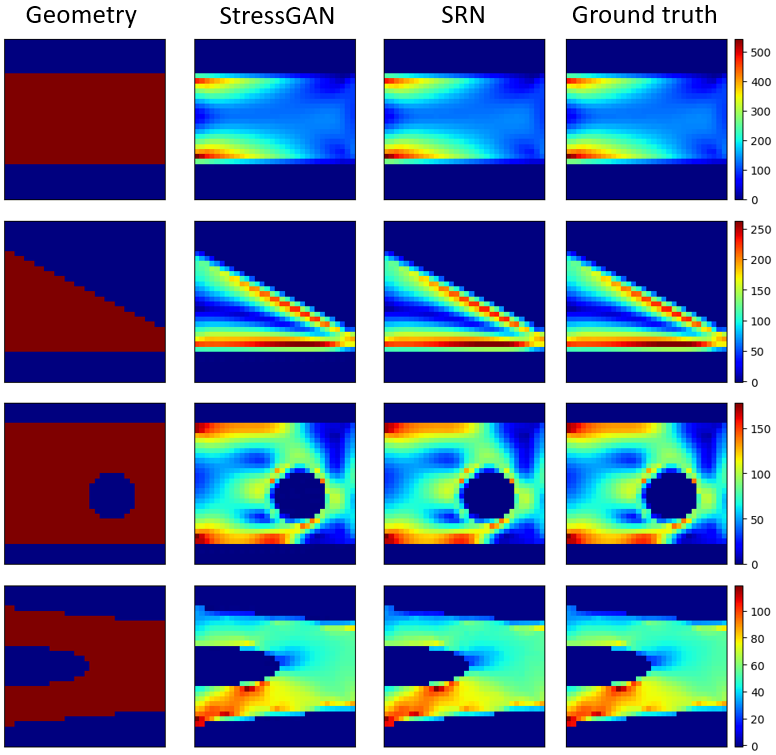}
	\caption{Evaluation of StressGAN and SRN on coarse-mesh dataset. Four evaluation cases are shown by each row. The visualizations of results of StressGAN and SRN are identical to the ground truth stress distributions. From left to right: 1) geometry (red); 2) predictions of StressGAN; 3) predictions of SRN; 4) ground truth stress distributions. The load magnitudes of each case: 
	1) $\mathrm{q}(x) = 27.5 N/mm^2$, $\mathrm{q}(y) = -47.6 N/mm^2$;
	2) $\mathrm{q}(x) = -43.0 N/mm^2$, $\mathrm{q}(y) = 61.4 N/mm^2$;
	3) $\mathrm{q}(x) = -3.5 N/mm^2$, $\mathrm{q}(y) = 19.7 N/mm^2$;
	4) $\mathrm{q}(x) = -54.8 N/mm^2$, $\mathrm{q}(y) = -4.8 N/mm^2$. (Units: mm-MPa-N) 
	}
	\label{fig:32_final}
\end{figure}

We also train and test our model using the coarse-mesh dataset with a split of 80\% - 20\% of training and test dataset. The evaluation results of StressGAN and SRN are shown in Table \ref{tab: Cantilever_beam}. Developed initially for this dataset, SRN performs well in this experiment and attains a better evaluation performance than StressGAN. With four layers removed, StressGAN still achieves a reasonably low error rate with a PMAE of $0.5\%$. This error rate is  low enough that it is difficult to tell the differences between the predicted results of the two networks through visualizations in Figure \ref{fig:32_final}.

\subsection{Generalization evalution}

\begin{figure}[ht]
	\centering
	\includegraphics[width=0.45\linewidth]{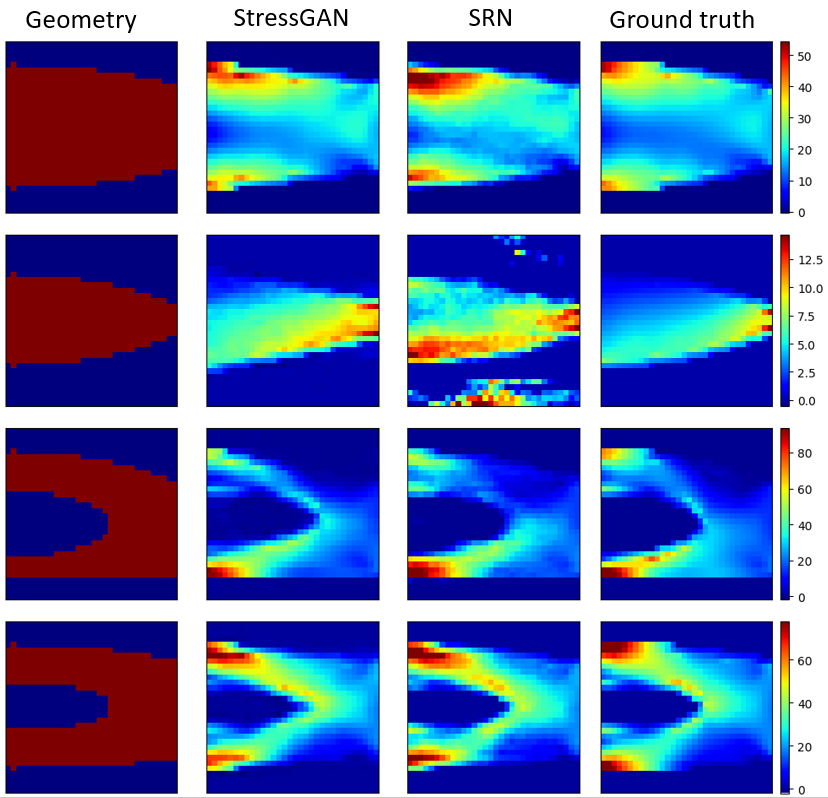}
	\caption{Evaluation of StressGAN and SRN on cases of different contours. Four evaluation cases are shown by each row. From left to right: 1) geometry (red); 2) predictions of StressGAN; 3) predictions of SRN; 4) ground truth stress distributions. 
	The load magnitudes of each case:
	1) $\mathrm{q}(x) = -5.7 N/mm^2$, $\mathrm{q}(y) = -8.2 N/mm^2$;
	2) $\mathrm{q}(x) = 10.0 N/mm^2$, $\mathrm{q}(y) = -0.9 N/mm^2$;
	3) $\mathrm{q}(x) = -7.7 N/mm^2$, $\mathrm{q}(y) = 6.4 N/mm^2$;
	4) $\mathrm{q}(x) = 5.0 N/mm^2$, $\mathrm{q}(y) = 8.7 N/mm^2$. (Units: mm-MPa-N)
	}
	\label{fig:cc2}
\end{figure}

\begin{figure}[H]
	\centering
	\includegraphics[width=0.45\linewidth]{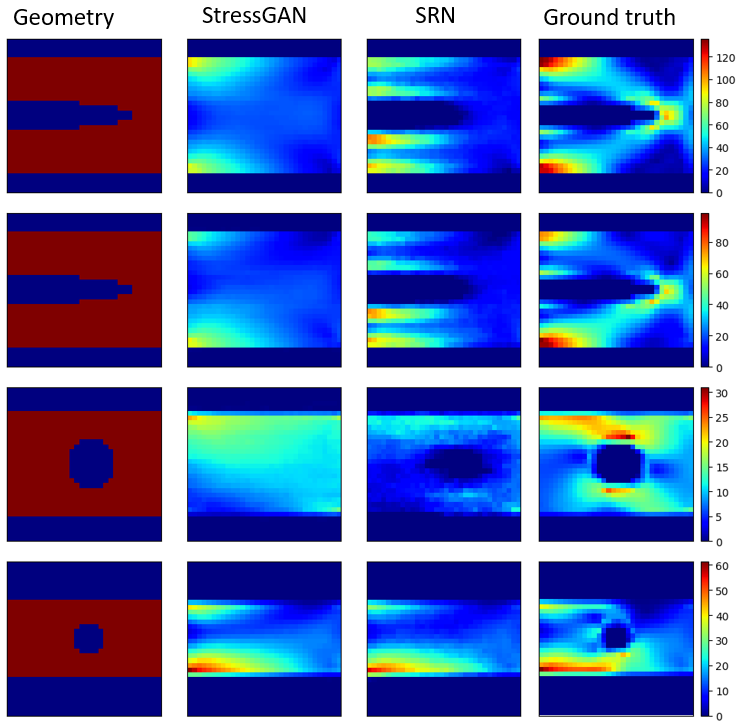}
	\caption{Evaluation of StressGAN and SRN on cases of cantilever beams with cellular openings. Models are trained with cantilever beams with openings and tested with cantilever beams without openings. Four evaluation cases are shown by each row. From left to right: 1) geometry (red); 2) predictions of StressGAN; 3) predictions of SRN; 4) ground truth stress distributions. 
	The load magnitudes of each case:
	1) $\mathrm{q}(x) = 1.7 N/mm^2$, $\mathrm{q}(y) = 9.8 N/mm^2$;
	2) $\mathrm{q}(x) = -7.7 N/mm^2$, $\mathrm{q}(y) = 6.4 N/mm^2$;
	3) $\mathrm{q}(x) = 10.0 N/mm^2$, $\mathrm{q}(y) = 0.9 N/mm^2$;
	4) $\mathrm{q}(x) = -9.1 N/mm^2$, $\mathrm{q}(y) = 4.2 N/mm^2$. (Units: mm-MPa-N)
	}
	\label{fig:cc3}
\end{figure}

\begin{figure}[H]
	\centering
	\includegraphics[width=0.5\linewidth]{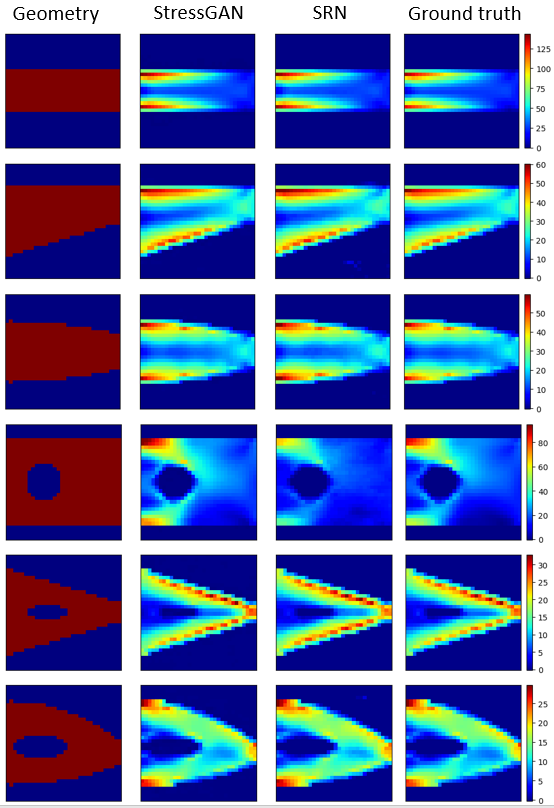}
	\caption{Evaluation of StressGAN and SRN on cases of different load orientations. This figure shows six evaluation cases of StressGAN and SRN when trained and tested with load conditions in different quadrants. From left to right: 1) geometry; 2) predictions of StressGAN; 3) predictions of SRN; 4) ground truth stress distributions. 
	The load magnitudes of each case:
	1) $\mathrm{q}(x) = -4.2 N/mm^2$, $\mathrm{q}(y) = -9.1 N/mm^2$;
	2) $\mathrm{q}(x) = -4.2 N/mm^2$, $\mathrm{q}(y) = -9.1 N/mm^2$;
	3) $\mathrm{q}(x) = -3.4 N/mm^2$, $\mathrm{q}(y) = -9.4 N/mm^2$;
	4) $\mathrm{q}(x) = -7.7 N/mm^2$, $\mathrm{q}(y) = -6.4 N/mm^2$. 
	5) $\mathrm{q}(x) = -3.4 N/mm^2$, $\mathrm{q}(y) = -9.4 N/mm^2$.
	6) $\mathrm{q}(x) = -4.2 N/mm^2$, $\mathrm{q}(y) = -9.1 N/mm^2$. (Units: mm-MPa-N)
	}
	\label{fig:angle}
\end{figure}

We conduct generalization experiments to explore our method's performance in  situations where the training dataset is sparse and testing data contains unseen cases. We include SRN and StressGAN into this experiment to compare their performances and demonstrate the characteristics of each network. The parametric results of the three experiments are shown in Table \ref{tab: first_sub}, \ref{tab: second_sub} and \ref{tab: third_sub}. The best performance under each metric is shown in bold. The selected samples of prediction results are shown in figure \ref{fig:cc2}, \ref{fig:cc3} and \ref{fig:angle} respectively. In general, StressGAN gives a better performance concerning the average prediction accuracy. The best PMAEs in three experiments are 8.23\%, 6.80\%, and 1.49\% respectively, which are all obtained by StressGAN. 

\begin{table}[!htb]
\centering
\caption{\label{tab: first_sub} Quantitative evaluation of StressGAN and SRN with training data of rectangular beams and trapezoidal beams and testing data of beams with a parabolar contour. The best performance under each metric is shown in bold. (Units: mm-MPa-N) }
%StressGAN achieves a better performance than SRN by all metrics. 
\begin{tabular}{cccccc}
\hline
Metric       & MSE                           & MAE                    & PMAE                           & PAE                   & PPAE                            \\ \hline
StressGAN & \textbf{28.91} & \textbf{2.80} & \textbf{7.50}\% & \textbf{6.85} & \textbf{18.10}\% \\ 
SRN       & 43.14                         & 3.28                         & 9.30\%                         & 12.55                        & 38.39\%                         \\ \hline
\end{tabular}
\end{table}

\begin{table}[!htb]
\centering
\caption{\label{tab: second_sub} Quantitative evaluation of StressGAN and SRN in the second sub-experiment with training data of beams without openings and testing data of beams with cellular openings. The best performance under each metric is shown in bold. (Units: mm-MPa-N) }
% The evaluation shows that this is a difficult task for both StressGAN and SRN. 

% StressGAN attains better MSE, MAE and PMAE results. SRN outperforms StressGAN in prediction of the largest stresses. 

\begin{tabular}{cccccc}
\hline
Metric   & MSE            & MAE     & PMAE            & PAE      & PPAE             \\ \hline
StressGAN & \textbf{77.20} & \textbf{4.40} & \textbf{6.80\%} & 16.96          & 24.10\%          \\
SRN       & 95.36          & 4.59          & 7.54\%          & \textbf{14.09} & \textbf{23.62\%} \\ \hline
\end{tabular}
\end{table}

\begin{table}[!htb]
\centering
\caption{\label{tab: third_sub} Quantitative evaluation of StressGAN and SRN in the experiment of cross-load direction training and evaluation. The best performance under each metric is shown in bold. (Units: mm-MPa-N)}
% The best performance under each metric is shown in bold. StressGAN outperforms SRN with a significant gap by all metrics. 
\begin{tabular}{cccccc}
\hline
Metric  & MSE            & MAE     & PMAE            & PAE      & PPAE             \\ \hline
StressGAN & \textbf{3.71} & \textbf{0.84} & \textbf{1.49\%} & \textbf{3.15}          & \textbf{4.86\% }         \\
SRN       & 6.86          & 1.29          & 2.58\%          & 6.72 & 11.66\% \\ \hline
\end{tabular}
\end{table}

In the two cross-geometry experiments, we can study the characteristics of StressGAN and SRN including their advantages and disadvantages.  Figure \ref{fig:cc2} shows the visualizations of the ground truth stress distributions and prediction stress distributions in the first cross-geometry experiment. Although the contour information of the input geometries is hard for StressGAN to capture, StressGAN outputs stress distributions closer to the samples in the dataset, especially in the regions of high stresses. Additionally, it generates a sharper (less blurred) prediction. Figure \ref{fig:cc3} shows  similar trends for the second cross-geometry experiment. On the one hand, StressGAN failed to predict stresses around the openings correctly. On the other hand, StressGAN generates more reasonable stress distributions which are more similar to the ground truth samples. Additionally, SRN could recognize the openings and predict zero stresses in void areas in some test cases. Since cellular openings have a considerable influence on stress concentrations and the networks have no explicit training on this phenomenon, large errors occur when we evaluate the predicted largest stress values as shown in Table \ref{tab: second_sub}.

The results of the cross-orientation experiment are shown in Figure \ref{fig:angle}. The output stress distributions from StressGAN are quite similar to the ground truths. From Table \ref{tab: third_sub}, it can be seen  that among the three generalization experiments, the cross-orientation experiment attains the best evaluation results. Since we use two images to express the load positions and magnitudes along the horizontal and vertical directions respectively, the deep learning method has a potential to learn the influence of the horizontal and vertical loads from the training dataset separately (essentially, the principle of superposition by exploting the linear nature of FEA) and synthesize reasonable results when tested on unseen load orientations. This  is especially useful in compressing the size of the training dataset for data efficiency without significantly increasing the error rate.

%%%%%%%%%%%%%%%%%%%%%%%%%%%%%%%%%%%%%%%%%%%%%%%%%%%%%%%
\section{Conclusions}
In this work, we develop a conditional generative adversarial network we call StressGAN for von Mises stress distribution prediction. StressGAN learns to predict the stress distribution given the geometries, load, and boundary conditions through a 2-player minimax game between its generator and discriminator. A fine-mesh stress distribution dataset composed of 38,400 cases of various geometries, load, and boundary conditions is proposed for evaluating the network's performance of complex stress prediction cases.

StressGAN achieves high accuracy in both experiments under multiple metrics, in evaluations of two stress distribution datasets. StressGAN outperforms the baseline model in both qualitative and quantitative evaluations in predicting the stress distributions given complicated geometry, displacement,  and load boundary conditions. It achieves an average error rate less than $0.21 \%$ on all stress values and $1.47 \%$ on the maximum stress value.
 
Moreover, StressGAN's performance under general scenarios is studied. StressGAN generates stress distributions more similar to samples in the dataset which shows it is a more effective learner in capturing the underlying knowledge of ground truth stress distributions. Furthermore, StressGAN is more efficient when facing unseen conditions. Although some  cases that lead to stress concentration such as  holes in geometries result in inaccurate predictions from StressGAN, the computed stress distributions still embody useful information such as the location of the highly stressed regions. The stress distributions are more similar to ground truths compared to the baseline method regardless of the conditions.
In contrast, our baseline model SRN is better at correctly estimating zero stresses in void areas but produces overall less accurate stress distributions under the same problem inputs. Furthermore, both StressGAN and SRN perform well given unseen load orientations compared to the cases where unseen geometries are involved.

In this work, the potential of generalizing the stress prediction ability to different categories is shown in generalization experiments. These findings constitute a one step toward generating data-driven analysis approaches that can generalize well to previously unseen problem configurations.  
%%%%%%%%%%%%%%%%%%%%%%%%%%%%%%%%%%%%%%%%%%%%%%%%%%%%%%%
\newpage

%% for ASME Conference and Journal
\bibliographystyle{unsrt}
% \bibliography{asme2e.bib}

%% for arXiv
%% 1. After compile by overleaf, select logs and output files, generate bbl files (such as manuscript.bbl). Or run LatexMk to generate a bbl file.
%% 2.Upload output.bbl file to the arXiv, add \input{references.bbl} at the end of the file before \end{docuement}
% \bibliographystyle{unsrt}

% \bibliography{references}

\begin{thebibliography}{10}

\bibitem{VonMisesTOPO}
RJ~Yang and CJ~Chen.
\newblock Stress-based topology optimization.
\newblock {\em Structural optimization}, 12(2-3):98--105, 1996.

\bibitem{miseschem}
D~Wang, J~Lee, K~Holland, T~Bibby, S~Beaudoin, and T~Cale.
\newblock Von mises stress in chemical-mechanical polishing processes.
\newblock {\em Journal of the Electrochemical Society}, 144(3):1121--1127,
  1997.

\bibitem{misesbone}
J~Chen, U~Akyuz, L~Xu, and RMV Pidaparti.
\newblock Stress analysis of the human temporomandibular joint.
\newblock {\em Medical Engineering \& Physics}, 20(8):565--572, 1998.

\bibitem{appliedFEA}
Larry~J Segerlind.
\newblock {\em Applied finite element analysis}, volume 316.
\newblock Wiley New York, 1976.

\bibitem{FEAconcepts}
Robert~D Cook et~al.
\newblock {\em Concepts and applications of finite element analysis}.
\newblock John Wiley \& Sons, 2007.

\bibitem{SF_DL}
Zhenguo Nie, Haoliang Jiang, and Levent~Burak Kara.
\newblock {Stress Field Prediction in Cantilevered Structures Using
  Convolutional Neural Networks}.
\newblock {\em Journal of Computing and Information Science in Engineering},
  20(1), 09 2019.
\newblock 011002.

\bibitem{Mises1913}
R.~v. Mises.
\newblock Mechanik der festen körper im plastisch- deformablen zustand.
\newblock {\em Nachrichten von der Gesellschaft der Wissenschaften zu
  Göttingen, Mathematisch-Physikalische Klasse}, 1913:582--592, 1913.

\bibitem{deeplearning}
Yann LeCun, Yoshua Bengio, and Geoffrey Hinton.
\newblock Deep learning.
\newblock {\em nature}, 521(7553):436, 2015.

\bibitem{surrogate_FEA}
Liang Liang, Minliang Liu, Caitlin Martin, and Wei Sun.
\newblock A deep learning approach to estimate stress distribution: a fast and
  accurate surrogate of finite-element analysis.
\newblock {\em Journal of The Royal Society Interface}, 15, 01 2018.

\bibitem{3Dtrusses}
Mehdi Nourbakhsh, Javier Irizarry, and John Haymaker.
\newblock Generalizable surrogate model features to approximate stress in 3d
  trusses.
\newblock {\em Engineering Applications of Artificial Intelligence}, 71:15--27,
  2018.

\bibitem{3Dprint}
Aditya Khadilkar, Jun Wang, and Rahul Rai.
\newblock Deep learning--based stress prediction for bottom-up sla 3d printing
  process.
\newblock {\em The International Journal of Advanced Manufacturing Technology},
  pages 1--15, 2019.

\bibitem{residual_pipe}
Jino Mathew, RJ~Moat, S~Paddea, Michael~E Fitzpatrick, and PJ~Bouchard.
\newblock Prediction of residual stresses in girth welded pipes using an
  artificial neural network approach.
\newblock {\em International Journal of Pressure Vessels and Piping},
  150:89--95, 2017.

\bibitem{shear}
Zohreh~Sheikh Khozani, Hossein Bonakdari, and Amir~Hossein Zaji.
\newblock Estimating the shear stress distribution in circular channels based
  on the randomized neural network technique.
\newblock {\em Applied Soft Computing}, 58:441--448, 2017.

\bibitem{FEA_modelUpdate1}
R.I. Levin and N.A.J. Lieven.
\newblock Dynamic finite element model updating using neural networks.
\newblock {\em Journal of Sound and Vibration}, 210(5):593 -- 607, 1998.

\bibitem{FEA_modelUpdate2}
M.J. Atalla and D.J. Inman.
\newblock On model updating using neural networks.
\newblock {\em Mechanical Systems and Signal Processing}, 12(1):135 -- 161,
  1998.

\bibitem{NN_plausibilityCheck}
Tobias Spruegel, Tina Schr{\"o}ppel, Sandro Wartzack, et~al.
\newblock Generic approach to plausibility checks for structural mechanics with
  deep learning.
\newblock In {\em DS 87-1 Proceedings of the 21st International Conference on
  Engineering Design (ICED 17) Vol 1: Resource Sensitive Design, Design
  Research Applications and Case Studies, Vancouver, Canada, 21-25.08. 2017},
  pages 299--308, 2017.

\bibitem{NN_FAE}
A.A. Javadi and T~P.~Tan.
\newblock Neural network for constitutive modelling in flnite element analysis.
\newblock {\em Computer Assisted Mechanics and Engineering Sciences}, 10, 01
  2003.

\bibitem{kmatrix}
Atsuya Oishi and Genki Yagawa.
\newblock Computational mechanics enhanced by deep learning.
\newblock {\em Computer Methods in Applied Mechanics and Engineering}, 327:327
  -- 351, 2017.
\newblock Advances in Computational Mechanics and Scientific Computation—the
  Cutting Edge.

\bibitem{PointNet}
R.~Qi Charles, Hao Su, Mo~Kaichun, and Leonidas~J. Guibas.
\newblock Pointnet: Deep learning on point sets for 3d classification and
  segmentation.
\newblock {\em 2017 IEEE Conference on Computer Vision and Pattern Recognition
  (CVPR)}, Jul 2017.

\bibitem{vessel}
Ali Madani, Ahmed Bakhaty, Jiwon Kim, Yara Mubarak, and Mohammad Mofrad.
\newblock Bridging finite element and machine learning modeling: stress
  prediction of arterial walls in atherosclerosis.
\newblock {\em Journal of biomechanical engineering}, 2019.

\bibitem{resnet}
Kaiming He, Xiangyu Zhang, Shaoqing Ren, and Jian Sun.
\newblock Deep residual learning for image recognition.
\newblock In {\em Proceedings of the IEEE conference on computer vision and
  pattern recognition}, pages 770--778, 2016.

\bibitem{senet}
Jie Hu, Li~Shen, and Gang Sun.
\newblock Squeeze-and-excitation networks.
\newblock In {\em Proceedings of the IEEE conference on computer vision and
  pattern recognition}, pages 7132--7141, 2018.

\bibitem{GAN}
Ian Goodfellow, Jean Pouget-Abadie, Mehdi Mirza, Bing Xu, David Warde-Farley,
  Sherjil Ozair, Aaron Courville, and Yoshua Bengio.
\newblock Generative adversarial nets.
\newblock In Z.~Ghahramani, M.~Welling, C.~Cortes, N.~D. Lawrence, and K.~Q.
  Weinberger, editors, {\em Advances in Neural Information Processing Systems
  27}, pages 2672--2680. Curran Associates, Inc., 2014.

\bibitem{goodfellow2016nips}
Ian Goodfellow.
\newblock Nips 2016 tutorial: Generative adversarial networks.
\newblock {\em arXiv preprint arXiv:1701.00160}, 2016.

\bibitem{pix2pix}
Phillip Isola, Jun{-}Yan Zhu, Tinghui Zhou, and Alexei~A. Efros.
\newblock Image-to-image translation with conditional adversarial networks.
\newblock {\em CoRR}, abs/1611.07004, 2016.

\bibitem{SRGAN}
Christian Ledig, Lucas Theis, Ferenc Husz{\'a}r, Jose Caballero, Andrew
  Cunningham, Alejandro Acosta, Andrew Aitken, Alykhan Tejani, Johannes Totz,
  Zehan Wang, et~al.
\newblock Photo-realistic single image super-resolution using a generative
  adversarial network.
\newblock In {\em Proceedings of the IEEE conference on computer vision and
  pattern recognition}, pages 4681--4690, 2017.

\bibitem{liu2017unsupervised}
Ming-Yu Liu, Thomas Breuel, and Jan Kautz.
\newblock Unsupervised image-to-image translation networks.
\newblock In {\em Advances in Neural Information Processing Systems}, pages
  700--708, 2017.

\bibitem{ROCGAN}
Grigorios~G Chrysos, Jean Kossaifi, and Stefanos Zafeiriou.
\newblock Robust conditional generative adversarial networks.
\newblock {\em arXiv preprint arXiv:1805.08657}, 2018.

\bibitem{DCGAN}
Alec Radford, Luke Metz, and Soumith Chintala.
\newblock Unsupervised representation learning with deep convolutional
  generative adversarial networks.
\newblock {\em CoRR}, abs/1511.06434, 2015.

\bibitem{ACN}
Jost~Tobias Springenberg, Alexey Dosovitskiy, Thomas Brox, and Martin~A.
  Riedmiller.
\newblock Striving for simplicity: The all convolutional net.
\newblock {\em CoRR}, abs/1412.6806, 2014.

\bibitem{Amir_phaseSegregatoin}
Amir~Barati Farimani, Joseph Gomes, Rishi Sharma, Franklin~L. Lee, and Vijay~S.
  Pande.
\newblock Deep learning phase segregation.
\newblock {\em CoRR}, abs/1803.08993, 2018.

\bibitem{Amir_Transport}
Amir~Barati Farimani, Joseph Gomes, and Vijay~S. Pande.
\newblock Deep learning the physics of transport phenomena.
\newblock {\em CoRR}, abs/1709.02432, 2017.

\bibitem{Lee_cylinder}
Sangseung {Lee} and Donghyun {You}.
\newblock {Data-driven prediction of unsteady flow fields over a circular
  cylinder using deep learning}.
\newblock {\em arXiv e-prints}, page arXiv:1804.06076, April 2018.

\bibitem{colagan}
Michela Paganini, Luke de~Oliveira, and Benjamin Nachman.
\newblock Calogan: Simulating 3d high energy particle showers in multilayer
  electromagnetic calorimeters with generative adversarial networks.
\newblock {\em Physical Review D}, 97(1):014021, 2018.

\bibitem{cloudRemove}
Kenji Enomoto, Ken Sakurada, Weimin Wang, Hiroshi Fukui, Masashi Matsuoka,
  Ryosuke Nakamura, and Nobuo Kawaguchi.
\newblock Filmy cloud removal on satellite imagery with multispectral
  conditional generative adversarial nets.
\newblock {\em CoRR}, abs/1710.04835, 2017.

\bibitem{ETH_galaxy}
Kevin Schawinski, Ce~Zhang, Hantian Zhang, Lucas Fowler, and Gokula~Krishnan
  Santhanam.
\newblock Generative adversarial networks recover features in astrophysical
  images of galaxies beyond the deconvolution limit.
\newblock {\em Monthly Notices of the Royal Astronomical Society: Letters},
  467(1):L110--L114, 2017.

\bibitem{CMU_dackEnergy}
Siamak {Ravanbakhsh}, Francois {Lanusse}, Rachel {Mandelbaum}, Jeff
  {Schneider}, and Barnabas {Poczos}.
\newblock {Enabling Dark Energy Science with Deep Generative Models of Galaxy
  Images}.
\newblock {\em arXiv e-prints}, page arXiv:1609.05796, September 2016.

\bibitem{CosmoGAN}
Mustafa {Mustafa}, Deborah {Bard}, Wahid {Bhimji}, Zarija {Luki{\'c}}, Rami
  {Al-Rfou}, and Jan {Kratochvil}.
\newblock {CosmoGAN: creating high-fidelity weak lensing convergence maps using
  Generative Adversarial Networks}.
\newblock {\em arXiv e-prints}, page arXiv:1706.02390, June 2017.

\bibitem{SRResnet}
Christian Ledig, Lucas Theis, Ferenc Huszar, Jose Caballero, Andrew~P. Aitken,
  Alykhan Tejani, Johannes Totz, Zehan Wang, and Wenzhe Shi.
\newblock Photo-realistic single image super-resolution using a generative
  adversarial network.
\newblock {\em CoRR}, abs/1609.04802, 2016.

\bibitem{solidspy}
Juan G{\'o}mez and N~Guar{\'\i}n-Zapata.
\newblock Solidspy: 2d-finite element analysis with python.”.
\newblock {\em Parameters}, 50:2--0, 2018.

\bibitem{Adam}
Diederik~P. Kingma and Jimmy Ba.
\newblock Adam: {A} method for stochastic optimization.
\newblock {\em CoRR}, abs/1412.6980, 2014.

\end{thebibliography}
\end{document}